\documentclass[10pt, conference]{IEEEtran}

\usepackage[utf8]{inputenc}

\usepackage[bookmarks=false]{hyperref}

\usepackage{lipsum}

\usepackage{caption}
\usepackage[table, xcdraw]{xcolor}
\usepackage{booktabs}

\usepackage{cite}

\usepackage{subimages}
\setfigdir{_figs}
\usepackage{tikz}
\usetikzlibrary{arrows} 
\usepackage{rotating}
\usepackage{caption}

\usepackage[cmex10]{amsmath}
\usepackage{amsthm}

\usepackage{amsmath}
\usepackage{amssymb}

\usepackage{algpseudocode}
\usepackage{algorithm}
\usepackage{multirow}

\usepackage{varwidth}

\usepackage{hyperref}

\usepackage{color}

\usepackage{booktabs}

\newcommand\copyrighttext{%
  \footnotesize \textcopyright 2016 IEEE. Personal use of this material is permitted.
  Permission from IEEE must be obtained for all other uses, in any current or future 
  media, including reprinting/republishing this material for advertising or promotional 
  purposes, creating new collective works, for resale or redistribution to servers or 
  lists, or reuse of any copyrighted component of this work in other works. 
  DOI: 10.1109/SIBGRAPI.2016.059}
\newcommand\copyrightnotice{%
\begin{tikzpicture}[remember picture,overlay]
\node[anchor=south,yshift=10pt] at (current page.south) {\fbox{\parbox{\dimexpr\textwidth-\fboxsep-\fboxrule\relax}{\copyrighttext}}};
\end{tikzpicture}%
}

\begin{document}



\title{Complexity-Aware Assignment of Latent Values in Discriminative Models for Accurate \\ Gesture Recognition}


\newif\iffinal
\finaltrue
\newcommand{\jemsid}{130}


\iffinal
  \author{%
    \IEEEauthorblockN{Manoel Horta Ribeiro, Bruno Teixeira, Ant\^onio Ot\'avio Fernandes, Wagner Meira Jr., Erickson R. Nascimento}
    \IEEEauthorblockA{%
      Computer Science Department\\
      Universidade Federal de Minas Gerais (UFMG), Brazil\\
      E-mail: {\tt\small \{manoelribeiro,bruno.texeira,otavio,meira,erickson\}@dcc.ufmg.br}
      }
  }
\else
  \author{Sibgrapi paper ID: \jemsid \\ }
\fi



\maketitle
\copyrightnotice

\begin{abstract}
Many of the state-of-the-art algorithms for gesture recognition are based on Conditional Random Fields (CRFs). 
Successful approaches, such as the Latent-Dynamic CRFs, extend the CRF by incorporating latent variables, whose values are mapped to the values of the labels.
In this paper we propose a novel methodology to set the latent values according to the gesture complexity.
We use an heuristic that iterates through the samples associated with each label value, estimating their complexity. We then use it to assign the latent values to the label values.
We evaluate our method on the task of recognizing human gestures from video streams. 
The experiments were performed in binary datasets, generated by grouping different labels.
Our results demonstrate that our approach outperforms the arbitrary one in many cases, increasing the accuracy by up to 10\%.
\end{abstract}

\IEEEpeerreviewmaketitle

\section{Introduction}
\label{it}
The fields of Computer Vision, Pattern Recognition and Human-Computer Interaction still face the challenging problems of gesture and activity recognition.  
Over the last two decades, generative models struggled to tackle these problems, and, eventually, with advancements in the inference methods for graphical models, Conditional Random Fields (CRFs)~\cite{crf} rose as a powerful discriminative alternative for dealing with them, relaxing the dependency assumptions on the inputs.

\begin{figure}[t]
\begin{center}
\includegraphics[width=0.95\linewidth]{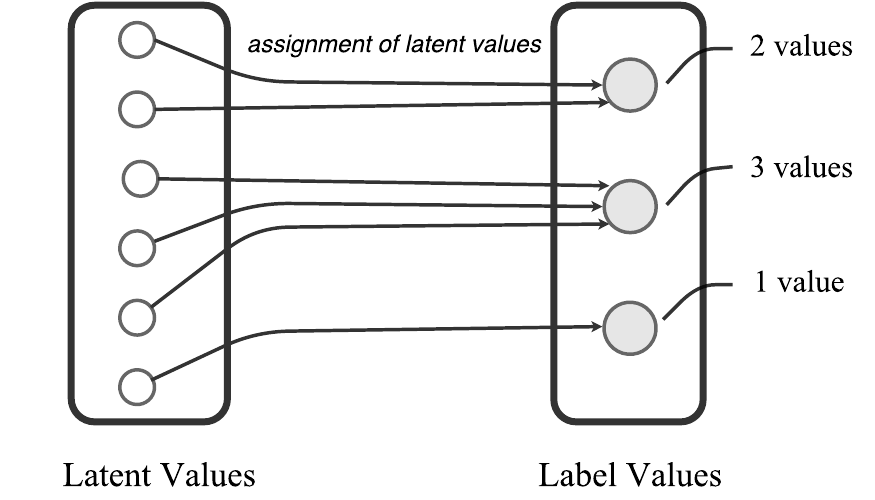}
\end{center}
  \captionof{figure}{Assignment of the latent values to the label values. The arrows show that each latent value maps into exactly one label value. We improve the existing models by determining this assignment  according to the complexity of the gestures. }
  \label{introfig}
\end{figure}

More recently, models based on CRFs have achieved the state-of-the-art results for such tasks~\cite{mvldcrf,hucrf,vicente2016wacv}. 
They are part of a bigger class of models which incorporate latent variables to the original CRF, increasing the model's expressiveness and its capacity to find relevant substructure in the gestures and activities~\cite{hcrf,ldcrf,mvldcrf,hucrf}. 
Two representative works of this approach are the Latent-Dynamic Conditional Random Fields (LDCRF)~\cite{ldcrf} and the Multi-View Latent-Dynamic Conditional Random Fields (MV-LDCRF)~\cite{mvldcrf}, where there is a direct mapping between the values of the latent variables and the label variables. In other words, there is a disjoint set of latent variable values associated with each label variable value~\footnote{Hereinafter we refer to latent variable values and label variable values as latent values and label values.}, as it can be seen in Figure \ref{introfig}. 

Although these methods improve the original CRF by training the latent variables as if they were the label variables, 
they add a new parameter that needs to be tuned: the number of latent values associated with each specific label value.
In this scenario, two questions arise: 
\textit{i)} how the assignment of these numbers impact the accuracy of the model, and if so,
\textit{ii)} how to systematically determine them.

Each latent value may be interpreted as a subset or a sub stage in a gesture or an activity. \textit{i.e.}
In the activity \textit{playing a sport}, using latent variables we may be able to model the difference between \textit{playing football} and \textit{playing chess}, which have a different set of gestures involved~\cite{hucrf}.
In a composite gesture, such as first raising the right hand, and then the left, we can use latent variables to model both stages differently, as well as their interaction~\cite{ldcrf}.
In this context, the assignment of the number of latent values plays a key role in the development of a model capable of representing the complexity of a gesture or an activity.

In this paper we propose an euclidean distance-based algorithm that, given the number of available latent values, determines their assignment among the labels.
Our approach inspects the samples associated with each label in order to get an associated complexity measurement, and then uses it to distribute the latent values.
We evaluate our method on the task of recognizing human gestures from unsegmented video streams in binary datasets using a LDCRF.
Our results show that the distribution of latent values has a high and non-trivial impact on the accuracy of the model, and that using our method  improves the performance by up to $10\%$ in accuracy in the tested datasets. 
The code of our method is available online\footnote{$https://github.com/manelhr/hidden\_states$}.

\section{Related Work}
\label{rw}
Many different probabilistic graphical models have been used for recognizing human gestures and activities, such as: Hidden Markov Models (HMMs)~\cite{hmmheadnod,hmmsignlan,hmmacti}, Dynamic Bayesian Networks (DBNs)~\cite{dinbay}, and Conditional Random Fields (CRFs)~\cite{crfaction,hucrf}. 
These graphical models are either generative or discriminative, depending on whether they make independence assumptions on the input or not.  As we usually rely on input that is heavily correlated, discriminative models tend to perform better~\cite{gdmodels}, especially for sequence labeling tasks~\cite{hucrf,ldcrf}. 
Sminchisescu~\textit{et al.}, for instance, outperformed a HMM approach using  a CRF for the task of classifying human motion activities, such as walking and jumping~\cite{crfaction}. 

In several approaches, there is an additional layer of latent variables, which are not observed, but inferred. This layer helps  to model complex labels~\cite{hucrf,ldcrf,hcrf,hu2l,mvldcrf} and is also used to exploit the underlying semantics between temporal segments~\cite{hucrf}.
Although the authors agree with the gains of using a layer of latent variables, they have diverging interpretations on how exactly they do so. 
Hu \textit{et al.} stated that latent variables represent a subset of a given label. For instance, take an hypothetical class\footnote{We use the terms class and label value interchangeably.} for the activity \textit{to open something}. The layer of latent variables would help to model the difference between \textit{opening a door} and \textit{opening a bottle}~\cite{hucrf}.
On the other hand, Morency \textit{et al.} claim that latent variables represent different stages of a same label being performed~\cite{ldcrf}. In a complex gesture or activity, different parts of the motion would have different label values, for example.

CRFs are discriminative models which were first introduced by Lafferty \textit{et al.}, and quickly adopted in the vision community~\cite{crf}. The two most successful extensions for gesture and activity recognition based on CRFs are the Hidden Conditional Random Fields (HCRFs) and the Latent-Dynamic Conditional Random Fields (LDCRFs).
The HCRFs add a layer of latent variables that are connected to a single label variable, which predicts the gesture or the activity which was performed for all the observations~\cite{hcrf}. 
The model presented by Hu~\textit{et al.} exploits, by using a HCRF-like structure, full connectivity between input, latent and label variables. The layer of latent variables added has a direct mapping to the label variables.
The LDCRFs models, beyond adding the layer of latent variables, have disjoint sets of latent values associated with each label value. They also include one label variable per observation, enabling the output to be a continuous stream, and outperforming models based on Support Vector Machines (SVMs), HMMs, CRFs and HCRFs~\cite{ldcrf}.
A representative LDCRFs approach is presented by Song \textit{et al.}~\cite{mvldcrf}. They proposed a multi-view version of the LDCRFs and also of the HCRFs. By splitting semantically related input variables in different views, the performance of these models was improved.

Hu~\textit{et al.} and Sung~\textit{et al.} also presented two-layered discriminative models for activity recognition. The model presented by Hu~\textit{et al.} recognizes sub-level activities and high-level activities successively, using the assumption that a high-level activity is composed of multiple sub-activities. While the first layer of the hierarchical model predicts low-level activity for each temporal segment, the second uses the sub-activity to estimate the high-level activity~\cite{hu2l}. They added a set of latent variables to enrich the expressiveness of the second layer, and use the sub-activity as an observed variable.
The model presented by Sung~\textit{et al.} learns the sub-activities implicitly, by considering them as latent variables~\cite{hmemm}.

In the aforementioned sequential discriminative models where the latent values have a direct mapping to the label, the assignment of the number of latent values per label value is chosen arbitrarily. Typically the same number of latent values is given for all label values~\cite{mvldcrf,hucrf,ldcrf}. Our work proposes a systematic way to distribute these, presenting significant performance gains in the tested datasets. 
Our method is valid for all the models of this category, but we mainly build upon the LDCRFs proposed by Morency \textit{et al.}~\cite{ldcrf}, since they were used to measure the performance improvements achieved by adopting our approach.

\section{Methodology}
\subsection{Theoretical Background}

In this section we briefly describe the main characteristics and modeling of Conditional Random Fields (CRFs) and Latent Dynamic Conditional Random Fields (LDCRFs). For details on how the inference or the parameter learning can be used, the reader is referred to~\cite{introductioncrfsutton,nowozin}. 

\subsubsection{Conditional Random Fields}


We review Conditional Random Fields as presented by Lafferty \textit{et al.} ~\cite{crf}, but using a factor graph notation for the sake of simplicity. 

Let $\mathbf{X}=(X_1,\ldots,X_k)$  be a set of observed random variables (e.g. feature vectors such as skeleton joints), and  $\mathbf{Y}=(Y_1,\ldots,Y_l)$ a set of target random variables (labels). $G$ is a factor graph $(V,E,A)$, whose nodes $V$ have a direct mapping with $\mathbf{X} \cup \mathbf{Y}$, and where $A$ is the set of factors associated with the edges $E$ that map the subset $\mathbf{X_a} \cup \mathbf{Y_a}$ into a positive real value\footnote{Notice that $\mathbf{X_a}\subset \mathbf{X},\mathbf{Y_a}\subset \mathbf{Y}$.}:

\begin{equation}
A=\{\psi_1,...,\psi_m\}, \psi_a: \mathbf{Y_a} \cup \mathbf{X_a} \mapsto \mathbb{R}^+. 
\end{equation}

The graph $G$ is a Conditional Random Field if $P(\mathbf{y}|\mathbf{x})$ factorizes according to Equation \ref{equationcrf}: 

\begin{equation}
\label{equationcrf} 
P(\mathbf{y}|\mathbf{x}) = 
\frac{1}{Z(\mathbf{x})}
\prod_{a \in A} \psi_a(\mathbf{y_a},\mathbf{x_a}),
\end{equation}

\begin{equation}
\label{normalization}
Z(\mathbf{x}) = \sum_{\mathbf{y} \in \mathbf{Y}}\prod_{a \in A} \psi_a(\mathbf{y_a},\mathbf{x_a}).
\end{equation}

\noindent where $Z(\mathbf{x})$ is the partition function. Lower case bold letters represent assignments to these sets of random variables, and $\mathbf{y_a}$ and $\mathbf{x_a}$ represent the values that correspond to the domain of the factor $\psi_a$.



\subsubsection{Linear Chain Conditional Random Fields}

\begin{figure}[t]
	\begin{center}
		\includegraphics[width=0.7\linewidth]{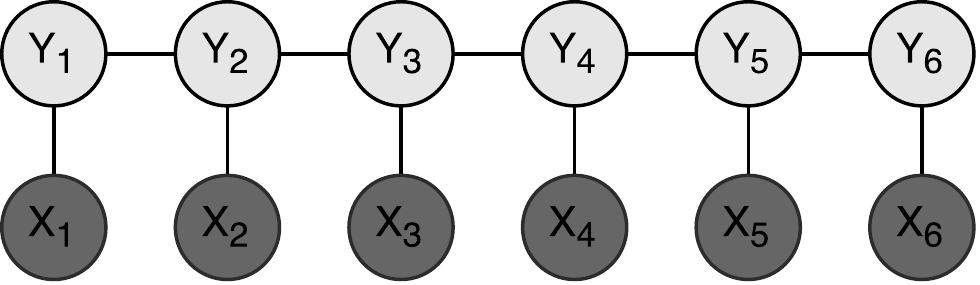}
	\end{center}
	\captionof{figure}{A CRF graphical model. The feature variables $X_i$ are connected directly to the label variables $Y_i$.}
	\label{crf-schema}
\end{figure}

Given the CRF model, is easy to derive a Linear Chain Conditional Random Field, which can be seen in Figure \ref{crf-schema}. For each edge $(i,j)$ in the chain, define a feature function:

\begin{equation}
f_k(v_i,v_j) = 
\begin{cases}
1 &\text{$(v_i,v_j) \in E$}\\
0 &\text{otherwise}
\end{cases}.
\end{equation}

Let $\log \psi_a$ be linear over the set of feature functions $K=\{f_{1},...,f_m\}$, and the vector of parameters of the model $\hat{\theta}=\{\theta_{1},...,\theta_m\}$:

\begin{equation}
\label{factor}
\psi_a(\mathbf{y_a},\mathbf{x_a}) = 
\exp\Bigg\{
\theta_{a} f_{a}(\mathbf{y_a},\mathbf{x_a})
\Bigg\}.
\end{equation}

We can then write the Linear Chain CRF as:

\begin{equation}
\label{equationlinearchaincrf} 
P(\mathbf{y}|\mathbf{x}) = 
\frac{1}{Z(\mathbf{x})}
\exp
\Bigg\{\sum_{t=1}^T \sum_{k=1}^K \theta_kf_k(y_t,y_{t-1},\mathbf{x_t})\Bigg\}.
\end{equation}

Notice that we iterate through all the $T$ levels of the chain and all the $K$ feature functions. We also modified the notation for feature functions so that they can represent the feature functions of both $label-label$ and $feature-label$ edges at once. The partition function $Z(\mathbf{x})$ is defined as:

\begin{equation}
\label{equationlinearchaincrffactor} 
Z(\mathbf{X}) = \sum_{{y_t,y_{t-1}} \in \mathbf{Y}}
\exp
\Bigg\{\sum_{t=1}^T \sum_{k=1}^K \theta_kf_k(y_t,y_{t-1},\mathbf{x_t})\Bigg\}.
\end{equation}


\begin{figure}[t]
\begin{center}
\includegraphics[width=0.7\linewidth]{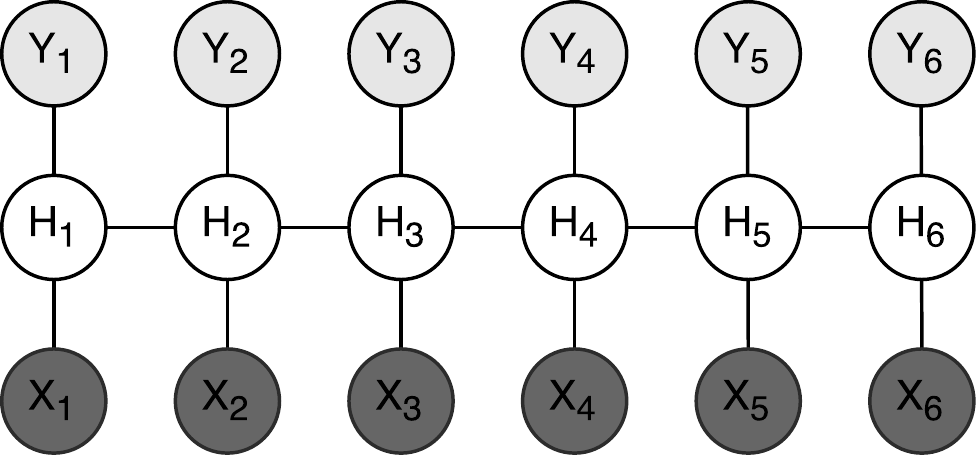}
\end{center}
  \captionof{figure}{A LDCRF graphical model. A new layer, composed of latent variables $H_i$, is used to enrich the expressiveness of the model. The latent
  	variables connect the label variables $Y_i$ to the feature variables $X_i$ and model the substructure of the gesture or activity.}
  \label{ldcrf-schema}
\end{figure}


\subsubsection{Latent-Dynamic Conditional Random Fields}
\label{bgldcrf}

Another extension for CRF models presented by Morency \textit{et al.}~\cite{ldcrf} includes a third set of random variables $ \mathbf{H} = (H_1, \ldots ,H_n)$. The values of variables $H_i$ have a direct mapping with the label values, and the probability of each label value a label variable $Y$ might assume is the sum of the probabilities of its associated latent values:

\begin{equation}
\label{ldcrf} 
P(\mathbf{y}|\mathbf{x}) = 
\sum_{\mathbf{\textbf{h}: \forall h_j \in \mathbf{H}_{Y_j}}}
P(\textbf{h}|\textbf{x}),
\end{equation}

\noindent where each $ h_j \in \mathbf{H}_{Y_j}$ is a member of the set of possible latent values for the class $Y_j$. The final probability distribution $P(\textbf{h}|\textbf{x})$ considering the latent variables instead of the label ones is given by:

\begin{equation}
\label{ldcrfeq} 
P(\mathbf{h}|\mathbf{x}) = 
\frac{1}{Z(\mathbf{x})}
\exp
\Bigg\{\sum_{t=1}^T \sum_{k=1}^K \theta_kf_k(h_t,h_{t-1},\mathbf{x})\Bigg\},
\end{equation}

\begin{equation}
\label{partitionldcrf} 
Z(\mathbf{x}) = \sum_{{h_t,h_{t-1}} \in \mathbf{H}}
\exp
\Bigg\{\sum_{t=1}^T \sum_{k=1}^K \theta_kf_k(h_t,h_{t-1},\mathbf{x})\Bigg\}.
\end{equation}

\noindent A graphical representation of a LDCRF model is depicted in Figure \ref{ldcrf-schema}.


\subsection{The Semantics of Latent Values}
\label{dhvv}

As stated, in a Latent Dynamic Conditional Random Field, we split each target variable value $y_t$ into a series of equivalent latent values $\mathbf{h}$. This increases the capacity of the model to represent complex labels, because it allows us to train a classifier to a larger number of labels than the existing ones~\cite{ldcrf}. 


Consider, for instance, an LDCRF that receives the positions of the joints of individuals as an input and has to decide which sports people are playing at each frame. There are a myriad of sports with  distinct gestures and actions involved, and, therefore, one could divide a class $sport$ into other simpler, more atomic classes, such as $(football,handball,chess)$. This is, intuitively, how Hu \textit{et al.} view the role of latent values~\cite{hucrf}. They allow different subsets of an activity or gesture to be represented by different label values, which improves the accuracy.

Another perspective is provided by Morency \textit{et al.}, which view the role of the latent variables as a way to model the sub stages of a given gesture or activity~\cite{ldcrf}. Let's consider a gesture with a complex substructure, such as a sequence of signals in American Sign Language (ASL). In this case, different latent values may model different parts of the gesture. Differently from Hu's perspective, the idea is that the same gesture may be modeled by different latent values in different stages of the gesture.

\begin{figure}[t]
\begin{center}
\includegraphics[width=0.70\linewidth]{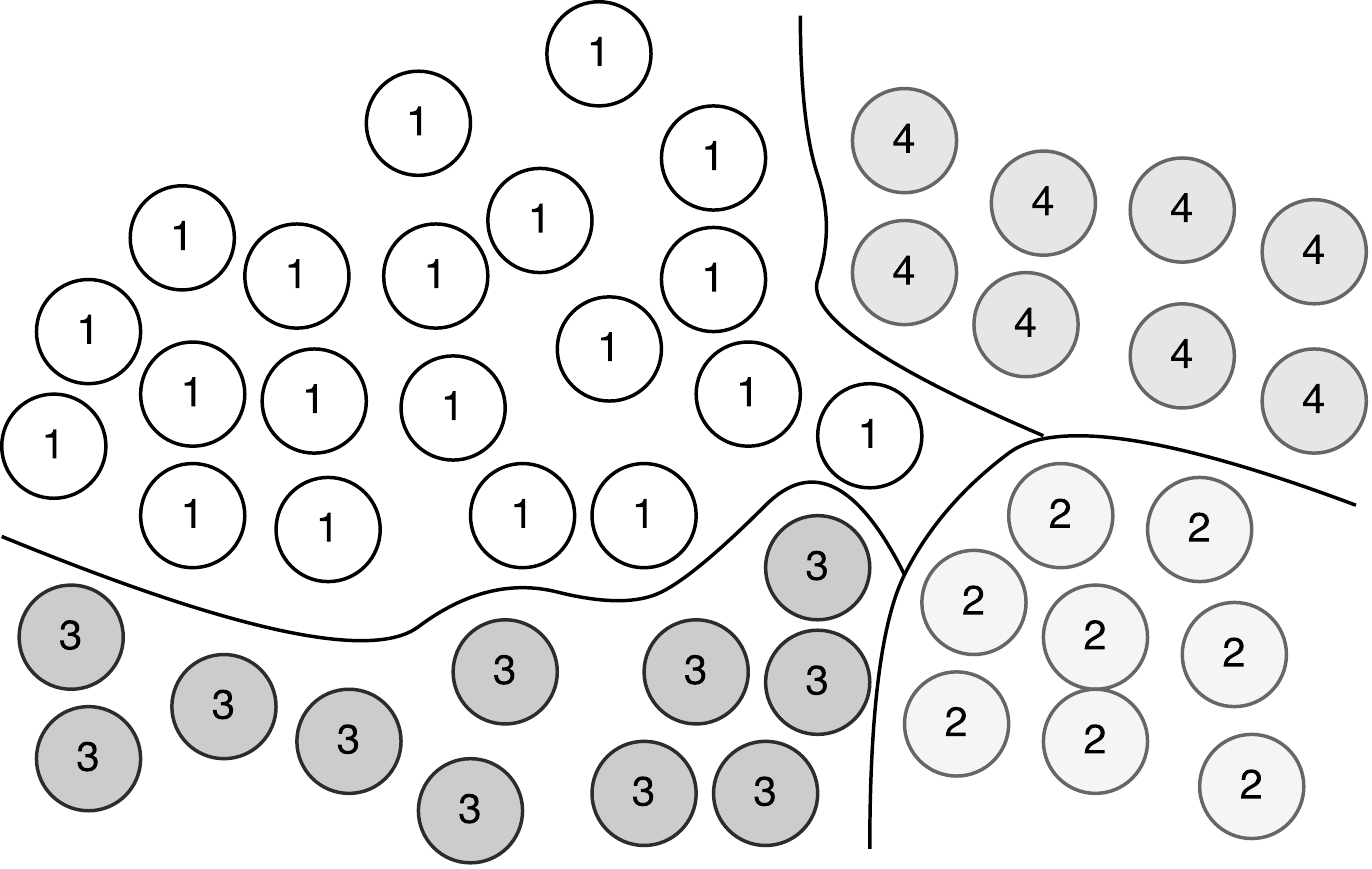}
\end{center}
  \captionof{figure}{Feature space of a CRF model. The lines separate the instances assigned to different label values.}
  \label{ap1}
\end{figure}
\begin{figure}[t]
\begin{center}
\includegraphics[width=0.70\linewidth]{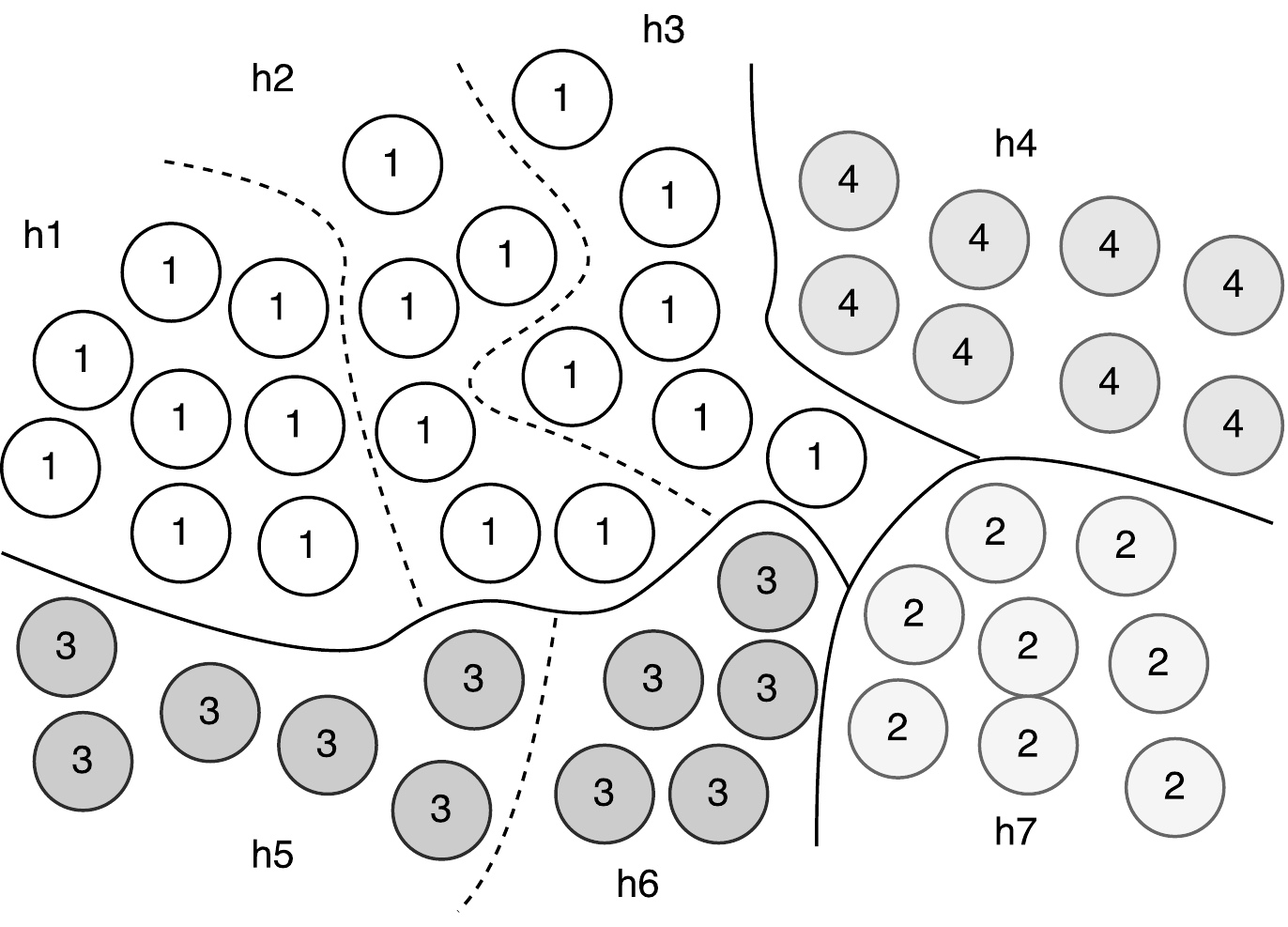}
\end{center}
  \captionof{figure}{Feature space of a latent model. Notice that we have $ \mathbf{H}_{Y_1}=\{h_1,h_2,h_3\},  \mathbf{H}_{Y_2}=\{h_7\},  \mathbf{H}_{Y_3}=\{h_5,h_6\}, \mathbf{H}_{Y_4}=\{h4\}$.  The dashed lines separate the instances assigned to different latent values. }
  \label{ap2}
\end{figure}

In the following sections, we describe our algorithm for assigning the number of latent values per class that takes those insights into consideration. Figures \ref{ap1} and \ref{ap2} illustrate the feature space of a CRF and of a LDCRF with different sets of latent values for each label, giving further insight on how this assignment is done. In Figure \ref{ap1} we classify the entries as one of the four labels directly, whereas in Figure \ref{ap2} we classify them as one of the seven values our latent variable might assume and then map those values into the label values.

\subsection{Computing The Complexities and Assigning Latent Values}


In order to decide which of the label values will benefit from having multiple latent values assigned to it, we propose a heuristic algorithm for finding the labels which can be better represented by many subsets or substages and the proportion to which they should receive latent values.

In the first step, we compute a measure of the complexity associated with each of the labels of the classifier according to Algorithm~\ref{measure}. 
Let $\mathcal{L}=\{l_1,l_2,l_3..., l_n\}$ be a set of labels in a dataset. We normalize the different samples that have the same label values~\footnote{Notice that each sample $(f_1,\ldots,f_n,l)$ is a tuple of time-series with the same length, where each $f_i$ is a numerical time-series which correspond to the values of the features and $l$ is a categorical time-series which correspond to the values of the labels.} (lines $2$ -- $8$) and then calculate the point-to-point euclidean distance between the pairs of features of these samples (lines $11$ -- $17$). This measures the complexities associated with each label, $v_0, \ldots, v_n$,  by estimating the difference among the samples of a given label. Each complexity measurement $v_i$ is then normalized, dividing it by the total sum of all measurements (lines $18$ -- $21$). This makes $v_{i} \in [0,1]$ and  $\sum_{\forall i \in Y} v_i = 1$. 

This heuristic is in agreement with the intuitions previously presented. If there are labels with different sub activities or sub gestures, as one would obtain from grouping \textit{opening a bottle} and \textit{opening a door} in the same label, when we calculate the point-to-point euclidean distance between the pairs of features, we would get a high value, as the samples from different sub activities will differ among themselves. Consequently, we then assign more latent values to these labels. 

On the other hand, if we have labels with a complex substructure, is it expected that there will be an overlap between different stages of the gesture when we take the point-to-point euclidean distance. Indeed, such assumption makes sense, since people will not execute each part of a complex gesture or activity in the exact same time. Notice that considering this setting we also will give a higher measure of complexity for labels with several sub stages.

Once we computed the complexities, we proceed to distribute the latent values according to Algorithm~\ref{dist}.
Our algorithm increments the number of latent values associated with a given label value such that the normalized array $norm$ will be close to the array $v$ calculated in the first algorithm (lines $5$ -- $11$). The algorithm receives as input the number of values we want to distribute, the complexity measures calculated in the previous step, the set of distinct labels and a real value $c\in[0,1]$. Notice that $c$  is a limit for the maximum percentage of latent values that one label value might claim. In the array $buckets=\{v_1,...,v_n\}$, the position $buckets[i]$ is the number of latent values assigned to the \textit{i-th} label.

\section{Experiments}
\label{experiments}

We evaluated our method on the task of unsegmented human gesture recognition using binary datasets. Those were created by grouping the labels of existing datasets such as the \textbf{ArmGesture} dataset and a subset of $6$ labels of the \textbf{NATOPS} dataset. Bellow we describe them, give details of our implementation and on the experiments protocol.

\subsection{Datasets}

\begin{figure*}[t]
	\centering
	\begin{tabular}{c}
		\includegraphics[width=0.85\linewidth]{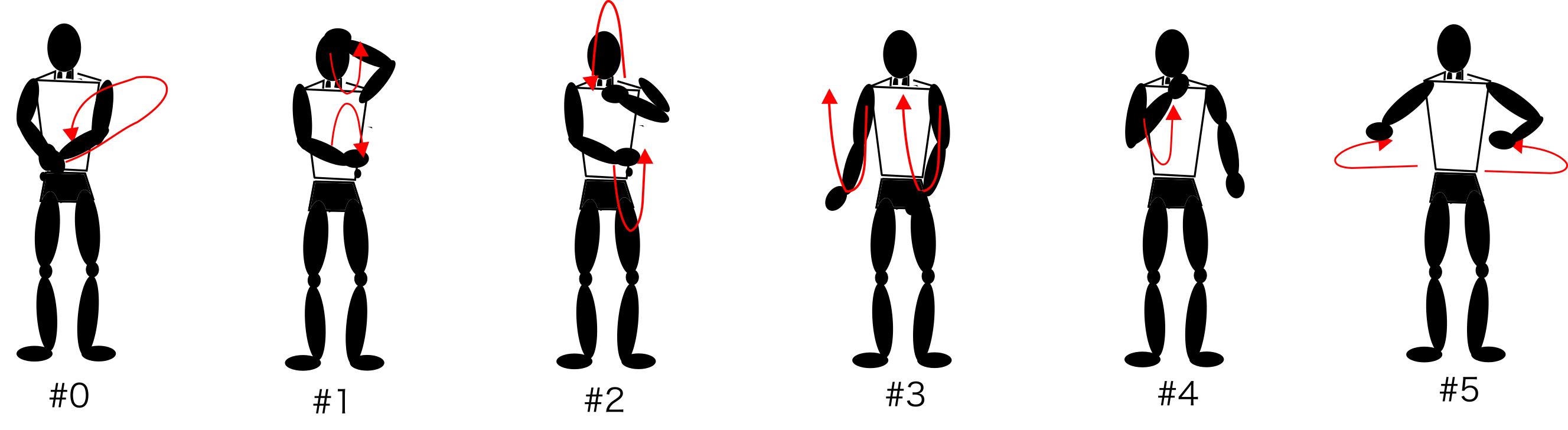} \\
		(a) Gestures from the \textbf{ArmGesture} dataset. \\
		\includegraphics[width=0.85\linewidth]{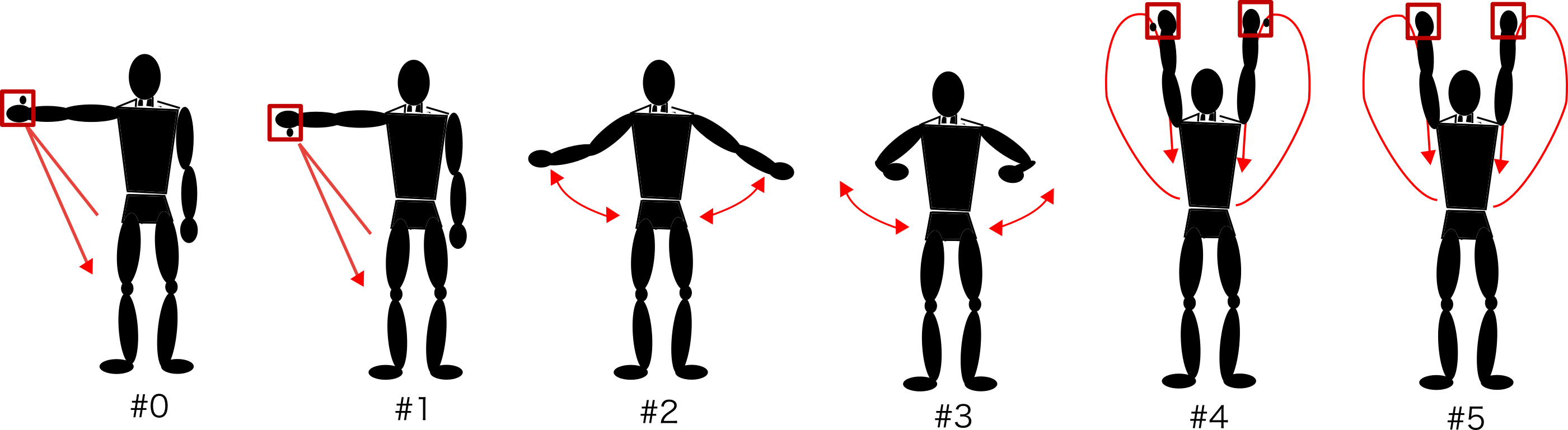} \\
		(b) Gestures from the \textbf{NATOPS} dataset. \\
	\end{tabular}
	\caption{Datasets used in experiments.}
	\label{ag_nt}
\end{figure*}

The \textbf{ArmGesture} dataset~\cite{quattoni2007hidden} contains the data on six arm gestures described by 2D joint angles and 3D euclidean coordinates for left/right shoulders and elbow. The six gestures are presented in Figure \ref{ag_nt} (a). The data was collected from $13$ participants with an average of $120$ samples per class, and subsampled by the factor of $2$~\cite{mvldcrf}. 

The \textbf{NATOPS} dataset \cite{song2011tracking} contains twenty-four body/hand gestures used when handling aircraft on the deck of an aircraft carrier. We used the same subset of gestures chosen by Song et al. \cite{mvldcrf}. These gestures can be seen in Figure \ref{ag_nt} (b).

One of drawbacks of these datasets is that the gestures do not differ  much in terms of complexity. Thus, in order to better measure the impact of our method, we generate datasets were different labels have different complexity measurements, but that are still based on real world data. Given the two original datasets, we make them binary by grouping the labels into two sets, each of which will become a new label. 
Notice that there are several grouping strategies. For instance, in a dataset with labels $(\{l_1\},\{l_2\},\{l_3\}...\{l_n\})$, one may group the label as $(\{l_1\},\{l_2,l_3...l_n\})$, $(\{l_1, l_2\},\{l_3...l_n\})$. 

Since we also want to evaluate how our model performs when the samples contain several gestures, we created new versions by grouping the different video segments of the original datasets in groups of three and concatenate them. We call these datasets \textit{many gestures} datasets, and the other ones \textit{single gesture} datasets.

Each created dataset is identified by the following expression:
$$  concat(id,\{1st\_labels\_set\},-,\{2nd\_labels\_set\}), $$

\noindent where $id$ is either \textit{AG} for the \textbf{ArmGesture} dataset or \textit{NT} for the \textbf{NATOPS} dataset, the label fields are the numbers of the labels grouped together (separated by an hyphen). 
The expression \textbf{NT01$-$2345}, for instance, refers to the dataset created by grouping the labels $\{0,1\}$ and the labels $\{2,3,4,5\}$ in the \textbf{NATOPS} dataset. We then have to specify if the dataset has \textit{many gestures} or a \textit{single gesture}.

\begin{algorithm}[t]
\begin{algorithmic}[1]
\Procedure{CompMeasure}{$instances$, $\mathcal{L}$}

\For{$l \in \mathcal{L}$}
\State $aux \gets$ instances of label $l$ 
\State $length\gets argmax(aux.length)$
\For{$\forall x \in aux$}
\State $x\gets linearInterpolation(x,length)$
\EndFor
\EndFor

\State $n \gets \mathcal{L}.length$
\State $v \gets zeros(n)$
\For{$l \in \mathcal{L}$}
\State $aux \gets$ instances of label $l$ 
\For{$\forall x,y \in aux$}
\State $dist \gets euclideanDistance(x,y)$ 
\State $v[l] \gets v[l] + dist$
\EndFor
\EndFor

\State $valuesSum \gets v[0] + ... + v[n]$
\For{$\forall s \in v$}
\State $s\gets s \div valuesSum$
\EndFor

\State \textbf{return} $v$
\EndProcedure
\end{algorithmic}
\caption{Algorithm for measuring the complexity of each label.}\label{measure}
\end{algorithm}

\begin{algorithm}[t]

\begin{algorithmic}[1]
\Procedure{Dist}{$number$, $values$, $\mathcal{L}$, $c$}

\State $n \gets \mathcal{L}.length$
\State $buckets \gets ones(n)$

\State $left = number - \mathcal{L}.length$
\While{$left \neq 0$}
\State $norm \gets (b_i) \div (\sum \limits_{j \in buckets} b_j), \forall i \in  buckets$
\State $dif \gets abs(buckets - values)$
\State $i \gets indexMin(dif)$ s.t. the normalized array where we we add $1$ to $b_i$ has $b_i < c$.
\State $left \gets left -1$
\State $bucket[i] \gets bucket[i] + 1$
\EndWhile
\State \textbf{return} $buckets$
\EndProcedure
\end{algorithmic}
\caption{Algorithm for distributing latent values across labels values given a complexity measurement.}\label{dist}
\end{algorithm}

\subsection{Experiments Protocol}
 
We performed three experiments with the datasets. The first one is a proof of concept on how the assignment of latent values may change the accuracy of a LDCRF. The second and the third ones make comparisons between the accuracy of a LDCRF whose states have been assigned using our method and arbitrarily.

\subsubsection{Experiment 1}

As stated, this experiment is a proof of concept to show that the assignment of the latent values is non-trivial and has a considerable impact in the accuracy. 

For a the \textbf{NT23$-$0145} \textit{many gestures} dataset we plot the confusion matrix (Figure~\ref{cm12}) of four different assignments of latent values.
In the first assignment, the first label value has two associated latent  values and the second one has only one, whereas in second assignment the first label value has one associated latent value and the second one has two. The remaining assignments are the cases where both labels values have either one and two latent values, respectively. 
The training and the test sets were obtained by dividing the dataset into three parts, from which $2/3rds$ were for the training and the remaining $1/3rd$ for testing.

\begin{table*}[!ht]
\centering
\caption{Results of our method in binary datasets. We present the accuracy and the standard deviation for two values of $c$ and the normal distribution. We also present the percentage of the complexity values calculated for each of the two binary labels in row \textit{CM}.}
\setlength\extrarowheight{1.5pt}
\resizebox{\linewidth}{!}{
\begin{tabular}{c|ccccc|cccc}
\hline  & & \multicolumn{4}{c}{Single Gesture} & \multicolumn{4}{c}{Many Gestures} \\ \cline{3-10}
\multirow{-2}{*}{Exp} & \multirow{-2}{*}{Dataset} & Ours c=1& Ours c=0.75 & Normal & CM & Ours c=1 & Ours c=0.75 & Normal  & CM  \\ \hline

 & \multicolumn{1}{c|}{AG0$-$12345} & $98.9  \pm 0.3$   & $97.2  \pm 3.3$ & $98.9  \pm 0.3$ & $02,98$ & $97.3  \pm 1.3$ & $97.5  \pm 0.9$ & $97.7  \pm 1.1$ & $03,97$\\
 & \multicolumn{1}{c|}{AG01$-$2345} & $99.2  \pm 0.4$   & $99.3  \pm 0.5$ & $99.2  \pm 0.4$ & $14,86$ & $85.0  \pm 24.8$ & $97.2  \pm 0.8$ & $97.3  \pm 1.3$ & $14,86$\\
 & \multicolumn{1}{c|}{AG012$-$345} & $94.3  \pm 0.4$   & $93.8  \pm 0.9$ & $94.3  \pm 0.4$ & $41,59$ & $89.2  \pm 2.9$ & $90.7  \pm 0.6$ & $90.6  \pm 0.9$ & $37,63$\\
 & \multicolumn{1}{c|}{NT0$-$12345} & $92.7  \pm 0.9$   & $92.8  \pm 1.0$ & $92.9  \pm 1.2$ & $02,98$ & $91.5  \pm 1.2$ & $92.6  \pm 0.8$ & $92.0  \pm 1.1$ & $03,97$\\
 & \multicolumn{1}{c|}{NT01$-$2345} & $92.2  \pm 1.6$   & $\mathbf{96.3  \pm 1.0}$ & $86.6  \pm 13.4$ & $13,87$ & $86.2  \pm 3.6$ & $\mathbf{93.8  \pm 0.4}$ & $83.8  \pm 22.0$ & $17,83$\\
\multirow{-6}{*}{II} & \multicolumn{1}{c|}{NT012$-$345} & $79.5  \pm 12.6$   & $73.4  \pm 12.6$ & $79.5  \pm 12.6$ & $45,55$ & $74.2  \pm 1.3$ & $\mathbf{79.5  \pm 17.6}$ & $74.2  \pm 1.3$ & $52,48$\\
\hline
 & \multicolumn{1}{c|}{AG01$-$2345}  & $99.2  \pm  0.4$ & $99.3  \pm  0.5$ & $99.2  \pm  0.4$ & $14,86$ & $85.0  \pm  24.8$ & $97.2  \pm  0.8$ & $97.3  \pm  1.3$ & $14,86$\\
 & \multicolumn{1}{c|}{AG12$-$0345}  & $95.6  \pm  1.0$ & $95.4  \pm  0.7$ & $95.4  \pm  0.7$ & $18,82$ & $90.1  \pm  1.6$ & $90.4  \pm  1.3$ & $90.5  \pm  1.9$ & $14,86$\\
 & \multicolumn{1}{c|}{AG23$-$0145}  & $99.8  \pm  0.3$ & $99.8  \pm  0.3$ & $99.8  \pm  0.3$ & $18,82$ & $96.9  \pm  1.7$ & $\mathbf{97.2  \pm  1.1}$ & $94.7  \pm  4.8$ & $19,81$ \\
 & \multicolumn{1}{c|}{AG34$-$0125}  & $96.1  \pm  1.3$ & $96.1  \pm  1.3$ & $96.1  \pm  1.3$ & $32,68$ & $90.8  \pm  1.6$ & $90.8  \pm  1.6$ & $90.8  \pm  1.6$ & $38,62$\\
 & \multicolumn{1}{c|}{AG45$-$0123}  & $98.2  \pm  1.0$ & $98.2  \pm  1.0$ & $98.2  \pm  1.0$ & $22,78$ & $\mathbf{96.3  \pm  1.2}$ & $95.4  \pm  0.7$ & $94.6  \pm  0.3$ & $27,73$ \\
 & \multicolumn{1}{c|}{NT01$-$2345}  & $92.2  \pm  1.6$ & $\mathbf{96.3  \pm  1.0}$ & $86.6  \pm  13.4$ & $13,87$ & $86.2  \pm  3.6$ & $\mathbf{93.8  \pm  0.4}$ & $83.8  \pm  22.0$ & $17,83$ \\
 & \multicolumn{1}{c|}{NT12$-$0345}  & $71.9  \pm  3.1$ & $71.8  \pm  1.1$ & $\mathbf{73.4  \pm  1.7}$ & $19,81$ & $70.5  \pm  1.6$ & $72.4  \pm  3.4$ & $72.0  \pm  2.3$ & $23,77$\\
 & \multicolumn{1}{c|}{NT23$-$0145}  & $\mathbf{92.2  \pm  1.7}$ & $84.0  \pm  0.9$ & $86.2  \pm  5.1$ & $24,76$ & $\mathbf{89.7  \pm  1.8}$ & $81.1  \pm  1.4$ & $85.7  \pm  12.0$ & $23,77$ \\
 & \multicolumn{1}{c|}{NT34$-$0125}  & $69.6  \pm  0.9$ & $69.8  \pm  1.0$ & $70.0  \pm  1.4$ & $22,78$ & $69.3  \pm  1.6$ & $\mathbf{69.7  \pm  1.6}$ & $66.8  \pm  1.8$ & $20,80$ \\
\multirow{-10}{*}{III}  & \multicolumn{1}{c|}{NT45$-$0123}  & $99.3  \pm  0.0$ & $99.3  \pm  0.1$ & $99.3  \pm  0.1$ & $20,80$ & $94.4  \pm  1.0$ & $\mathbf{94.7  \pm  1.2}$ & $87.8  \pm  10.4$ & $21,79$ \\
\hline
\end{tabular}}
\label{t1}

\end{table*}

\subsubsection{Experiment 2 and 3}

For Experiments 2 and 3, we compare the accuracy of training a LDCRF with our method of assigning latent values (for two values of $c$, $0.75$ and $1$) with the accuracy of an arbitrary assignment. 
We split the data into a test set of $1/3rd$ of the data, and performed a 5-fold cross validation in the remaining $2/3rds$. 
We then obtain total number of latent values to which had the best performance in the 5-fold cross validation for our method of assignment and for the arbitrary one. This parameters is then used to train the model with what was the validation $2/3rds$ of the data and test it with the $1/3rd$ which was originally the test set. This procedure is repeated $3$ times, varying the validation set chosen initially (as in a 5-fold cross validation nested inside a 3-fold cross validation).

In the validation phase, we selected the total number of latent variable values from a predetermined number of possibilities. For the binary datasets created from the \textbf{ArmGesture} dataset those values were $\{2,4,6,8,10,12,14\}$ and for the datasets created from the \textbf{NATOPS} dataset those values were $\{2,4,6,8\}$.
The choice of these values adopted three criteria: (i) even numbers, so that the arbitrary assignment always gave all labels the same amount of latent values, (ii) feasibility for computing time, (iii) values that do not over-fit the model. 

In Experiment 2 we created the datasets using the following algorithm: we start with the label grouping $\{\}, \{l_0,l_1,l_2,l_3,l_4,l_5\}$.  Then we incrementally create new datasets by adding the label with the lowest index from the second group to the first group and generating the correspondent binary dataset. We stop when both groups have the same number of labels. 
The idea behind exploring this pattern is to see how the difference in the complexity of binary labels would affect the method. Notice that this is possible because as we add more label values to be collapsed in the first of the mentioned groups, we decrease the difference of the complexity measurement between the two label values.

In Experiment 3 we created the datasets by combining the labels in all possible forms to obtain two groups, containing two and three labels respectively. These labels are then collapsed to generate the binary labels of the new dataset. 
We explore this pattern to see how our assignment would improve the LDCRF given a more subtle difference in the  complexity measurement of the binary label values.

\subsection{Implementation Details}

We implemented our Latent Dynamic CRF based on the Latent Graph CRF model from \textit{PyStruct} \cite{pystruct}. For the inference we use the Quadratic Pseudo-Boolean Optimization (QPBO), with the interface also provided by \textit{PyStruct}. The initial assignment of the latent values was done with a deterministic implementation of the k-means clustering technique, which uses a PCA to initialize its centroids. 

\section{Results}
In this section we present and analyze the results of the experiments proposed in Section \ref{experiments}.

\subsection{Experiment 1}

\begin{figure}[t]
	\begin{center}
		\includegraphics[width=0.7\linewidth]{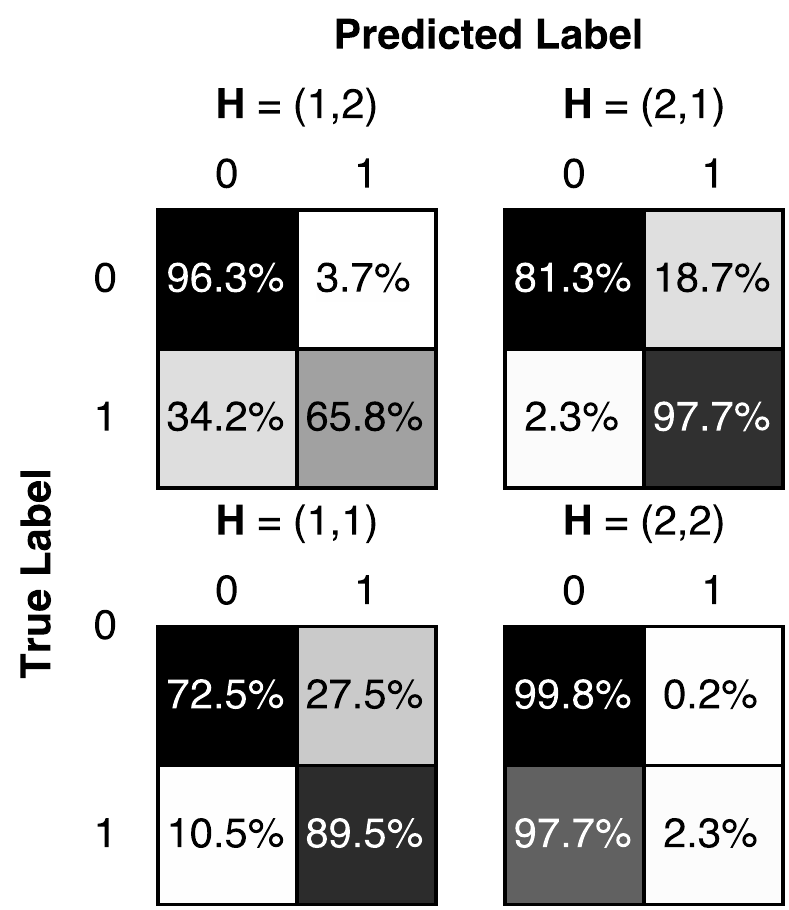}
	\end{center}
	\captionof{figure}{Confusion matrix for the \textbf{NT23$-$0145} \textit{many gestures} dataset using different assignments of latent values. The rows of the matrix are normalized, for comparing the accuracy of the classes and the shading of each cell is proportional to the number of instances that correspond to the given scenario.}
	\label{cm12}
\end{figure}

The confusion matrix of the assignment of latent values, $(1,2),(2,1),(1,1),(2,2)$~\footnote{Notice that in this case $(x,y)$ means that $(|\mathbf{H}_{Y_1}| = x, |\mathbf{H}_{Y_2}| = y)$.}  for the \textbf{NT23$-$0145} dataset, can be seen in Figure \ref{cm12}. Each row of each matrix is normalized so that we can compare the gains of a given latent value assignment for each of the classes.
The results show that the model with the $(2,1)$ assignment has the better accuracy trade-off,  with approximately  $89.5\%$  average accuracy \footnote{The average accuracy is calculated by multiplying the values shown in the table by the number of instances in each of the cases.}.
It is worth mentioning that the model with the $(2,2)$ assignment is significantly worse than all the others, with only $51.5\%$ average accuracy. A clear example of a case where  more latent values do not correspond to a better model.
Together, these results show that assigning the latent values according to our computed complexity measurement improves the accuracy of the LDCRF. Additionally, they show that this assignment has a non-trivial impact in the accuracy of the model.

\subsection{Experiment 2 and 3}

The results for Experiments 2 and 3 are shown in Table~\ref{t1}. We present results for our approach for two values of $c$ and also for the arbitrary assignment. The column \textit{CM} contains the calculated complexity measurement for the two labels in the dataset.
The cells in bold indicate cases where the accuracy achieved by training the model with our method was $1\%$ bigger than the arbitrary choice, as well as cases where the arbitrary choice outperformed ours by $1\%$. 
Notice that the second scenario happened once, in the \textbf{NT12$-$0345} dataset, where the arbitrary assignment had an accuracy slightly higher than ours. 
On the other hand, our method outperformed the arbitrary choice by significant values in the many cases. In the \textbf{NT01$-$2345} single gesture dataset and in the \textbf{NT01$-$2345} many gestures datasets, for instance, we increased the accuracy of the method by roughly $10\%$.

Two interesting questions that might arise given the results shown are \textit{(i)} why does our method gives much better results with the \textbf{NATOPS} dataset than with the \textbf{ArmGesture} dataset and \textit{(ii)} why does our method performs better in datasets where the difference in the estimated complexity for the binary labels is not too big. 
We draw the following observations. The low gains in performance in the \textbf{ArmGesture} dataset is probably due to how easy the dataset is. The accuracy is already too high with modeling that uses just a few latent values, and adding more models just makes the dataset overfit. In many cases the models selected in the validation for the ArmGesture dataset used only two latent values, which is equivalent to a normal linear chain CRF.
It is worth noting that, when we have a larger imbalance in the complexity of the binary labels, it also becomes easier to classify, as one must only learn less complex labels. This might be the reason of the better performance of our method compared to the arbitrary assignment in datasets whose difference in the associated complexity of the binary label is small. 
\section{Conclusion}
The assignment values of latent values is an important parameter in CRF based models with a layer of latent variables.  The previous work on such models had a very intuitive approach towards the purpose of the latent variables, and did not address the problem of finding an ideal assignment of these values throughout the label values.

We introduced a new method for assigning them in Latent-Dynamic Conditional Random Fields which outperforms an arbitrary distribution of those values. We evaluated our approach using real world binary datasets for the task of unsegmented human gesture recognition, and provided empirical results of the advantages of our method. The method is generalizable for all the models with a latent layer whose values have a direct mapping with the label values.

Our method showed significant improvements in the hardest of the datasets we used for benchmarking (the NATOPs dataset), and was particularly effective when the difference in the complexity measurement calculated was not too high.  By applying our technique we increased the recognition accuracy by roughly $10\%$ in some cases.

In the future, it is of interest to investigate the semantics of the latent layer and its impact on the models performance. To help in such investigation, we plan to create a new dataset where the labels have a significant difference with respect to complexity.

\iffinal
\section*{Acknowledgment}

The authors would like to thank the agencies CAPES, CNPq, FAPEMIG, InWeb, MasWeb and BigSea for funding different parts of this work. 
\fi


\bibliographystyle{IEEEtran}
\bibliography{paper}

\begin{thebibliography}{10}
\providecommand{\url}[1]{#1}
\csname url@samestyle\endcsname
\providecommand{\newblock}{\relax}
\providecommand{\bibinfo}[2]{#2}
\providecommand{\BIBentrySTDinterwordspacing}{\spaceskip=0pt\relax}
\providecommand{\BIBentryALTinterwordstretchfactor}{4}
\providecommand{\BIBentryALTinterwordspacing}{\spaceskip=\fontdimen2\font plus
\BIBentryALTinterwordstretchfactor\fontdimen3\font minus
  \fontdimen4\font\relax}
\providecommand{\BIBforeignlanguage}[2]{{%
\expandafter\ifx\csname l@#1\endcsname\relax
\typeout{** WARNING: IEEEtran.bst: No hyphenation pattern has been}%
\typeout{** loaded for the language `#1'. Using the pattern for}%
\typeout{** the default language instead.}%
\else
\language=\csname l@#1\endcsname
\fi
#2}}
\providecommand{\BIBdecl}{\relax}
\BIBdecl

\bibitem{crf}
J.~Lafferty, A.~McCallum, and F.~C. Pereira, ``Conditional random fields:
  Probabilistic models for segmenting and labeling sequence data,'' 2001.

\bibitem{mvldcrf}
Y.~Song, L.-P. Morency, and R.~Davis, ``Multi-view latent variable
  discriminative models for action recognition,'' in \emph{Computer Vision and
  Pattern Recognition (CVPR), 2012 IEEE Conference on}.\hskip 1em plus 0.5em
  minus 0.4em\relax IEEE, 2012, pp. 2120--2127.

\bibitem{hucrf}
N.~Hu, G.~Englebienne, Z.~Lou, and B.~Krose, ``Learning latent structure for
  activity recognition,'' in \emph{Robotics and Automation (ICRA), 2014 IEEE
  International Conference on}.\hskip 1em plus 0.5em minus 0.4em\relax IEEE,
  2014, pp. 1048--1053.

\bibitem{vicente2016wacv}
C.~M. de~Souza~Vicente, E.~R. Nascimento, L.~E.~C. Emery, C.~A.~G. Flor,
  T.~Vieira, and L.~B. Oliveira, ``High performance moves recognition and
  sequence segmentation based on key poses filtering,'' in \emph{{2016 IEEE
  Winter Conference on Applications of Computer Vision (WACV)}}, March 2016,
  pp. 1--8.

\bibitem{hcrf}
S.~B. Wang, A.~Quattoni, L.~Morency, D.~Demirdjian, and T.~Darrell, ``Hidden
  conditional random fields for gesture recognition,'' in \emph{Computer Vision
  and Pattern Recognition, 2006 IEEE Computer Society Conference on}, vol.~2,
  2006, pp. 1521--1527.

\bibitem{ldcrf}
L.~Morency, A.~Quattoni, and T.~Darrell, ``Latent-dynamic discriminative models
  for continuous gesture recognition,'' in \emph{Computer Vision and Pattern
  Recognition, 2007. CVPR '07. IEEE Conference on}, June 2007, pp. 1--8.

\bibitem{hmmheadnod}
S.~Fujie, Y.~Ejiri, K.~Nakajima, Y.~Matsusaka, and T.~Kobayashi, ``A
  conversation robot using head gesture recognition as para-linguistic
  information,'' in \emph{Robot and Human Interactive Communication, 2004.
  ROMAN 2004. 13th IEEE International Workshop on}.\hskip 1em plus 0.5em minus
  0.4em\relax IEEE, 2004, pp. 159--164.

\bibitem{hmmsignlan}
M.~Assan and K.~Grobel, ``Video-based sign language recognition using hidden
  markov models,'' in \emph{Gesture and Sign Language in Human-Computer
  Interaction}.\hskip 1em plus 0.5em minus 0.4em\relax Springer, 1997, pp.
  97--109.

\bibitem{hmmacti}
J.~Sung, C.~Ponce, B.~Selman, and A.~Saxena, ``Unstructured human activity
  detection from rgbd images,'' in \emph{Robotics and Automation (ICRA), 2012
  IEEE International Conference on}.\hskip 1em plus 0.5em minus 0.4em\relax
  IEEE, 2012, pp. 842--849.

\bibitem{dinbay}
Y.-c. Ho, C.-h. Lu, I.-h. Chen, S.-s. Huang, C.-y. Wang, L.-c. Fu
  \emph{et~al.}, ``Active-learning assisted self-reconfigurable activity
  recognition in a dynamic environment,'' in \emph{Proceedings of the 2009 IEEE
  international conference on Robotics and Automation}.\hskip 1em plus 0.5em
  minus 0.4em\relax IEEE Press, 2009, pp. 1567--1572.

\bibitem{crfaction}
C.~Sminchisescu, A.~Kanaujia, and D.~Metaxas, ``Conditional models for
  contextual human motion recognition,'' \emph{Computer Vision and Image
  Understanding}, vol. 104, no.~2, pp. 210--220, 2006.

\bibitem{gdmodels}
A.~Y. Ng and M.~I. Jordan, ``On discriminative vs. generative classifiers: A
  comparison of logistic regression and naive bayes,'' in \emph{Advances in
  Neural Information Processing Systems 14}, T.~G. Dietterich, S.~Becker, and
  Z.~Ghahramani, Eds.\hskip 1em plus 0.5em minus 0.4em\relax MIT Press, 2002,
  pp. 841--848.

\bibitem{hu2l}
N.~Hu, G.~Englebienne, and B.~Krose, ``A two-layered approach to recognize
  high-level human activities,'' in \emph{Robot and Human Interactive
  Communication, 2014 RO-MAN: The 23rd IEEE International Symposium on}, Aug
  2014, pp. 243--248.

\bibitem{hmemm}
J.~Sung, C.~Ponce, B.~Selman, and A.~Saxena, ``Unstructured human activity
  detection from rgbd images,'' in \emph{Robotics and Automation (ICRA), 2012
  IEEE International Conference on}, May 2012, pp. 842--849.

\bibitem{introductioncrfsutton}
C.~Sutton and A.~McCallum, ``An introduction to conditional random fields,''
  \emph{Machine Learning}, vol.~4, no.~4, pp. 267--373, 2011.

\bibitem{nowozin}
S.~Nowozin and C.~H. Lampert, ``Structured learning and prediction in computer
  vision,'' \emph{Found. Trends. Comput. Graph. Vis.}, vol.~6, pp. 185--365,
  Mar. 2011.

\bibitem{quattoni2007hidden}
A.~Quattoni, S.~Wang, L.-P. Morency, M.~Collins, and T.~Darrell, ``Hidden
  conditional random fields,'' \emph{IEEE Transactions on Pattern Analysis \&
  Machine Intelligence}, no.~10, pp. 1848--1852, 2007.

\bibitem{song2011tracking}
Y.~Song, D.~Demirdjian, and R.~Davis, ``Tracking body and hands for gesture
  recognition: Natops aircraft handling signals database,'' in \emph{Automatic
  Face \& Gesture Recognition and Workshops (FG 2011), 2011 IEEE International
  Conference on}.\hskip 1em plus 0.5em minus 0.4em\relax IEEE, 2011, pp.
  500--506.

\bibitem{pystruct}
A.~C. M{\"u}ller and S.~Behnke, ``pystruct - learning structured prediction in
  python,'' \emph{Journal of Machine Learning Research}, vol.~15, pp.
  2055--2060, 2014.

\end{thebibliography}

\end{document}